# Error Detection in Large-Scale Natural Language Understanding Systems Using Transformer Models

**Rakesh Chada, Pradeep Natarajan, Darshan Fofadiya, Prathap Ramachandra**
Amazon Alexa AI
{rakchada, natarap, darshanf, prathara}@amazon.com

## Abstract

Large-scale conversational assistants like Alexa, Siri, Cortana and Google Assistant process every utterance using multiple models for domain, intent and named entity recognition. Given the decoupled nature of model development and large traffic volumes, it is extremely difficult to identify utterances processed erroneously by such systems. We address this challenge to detect domain classification errors using offline Transformer models. We combine utterance encodings from a RoBERTa model with the N-best hypothesis produced by the production system. We then fine-tune end-to-end in a multi-task setting using a small dataset of human-annotated utterances with domain classification errors. We tested our approach for detecting misclassifications from one domain that accounts for <0.5% of the traffic in a large-scale conversational AI system. Our approach achieves an F1 score of 30% outperforming a bi-LSTM baseline by 16.9% and a standalone RoBERTa model by 4.8%. We improve this further by 2.2% to 32.2% by ensembling multiple models.

## 1   Introduction

Conversational assistants such as Cortana, Google Assistant, Alexa and Siri leverage multiple machine learning models and services like automatic speech recognition (ASR), natural language understanding (NLU), entity resolution (ER), utterance routing and text-to-speech (TTS). In particular, the NLU system is modularized into multiple domains such as Music, Movies, Weather etc. These domain teams then train one-vs-all domain, intent and named entity recognition models independently. At run time, a re-ranker is used to sort the output of these independent models and route the utterance to the right domain. This is a well-known design pattern used in large scale conversational assistants (Sarikaya, 2017; Su et al., 2018).

This modularized architecture allows scaling of NLU systems to support multiple domains catering to billions of diverse utterances. However, it is extremely challenging to isolate the source of error in utterances where the system fails, both due to the large traffic volume and the asynchronous nature of model development and updates by multiple independent teams. In particular, we focus on identifying utterances with domain classification errors, which we call false rejects (FR). Such errors can arise due to errors in individual domain classification models, re-ranker or routing algorithm and are extremely hard to isolate. To illustrate this challenge, consider a domain $X$ that accounts for 0.5% of the traffic in a conversational assistant receiving Y utterances every week. Also assume that 20% of the utterances belonging to $X$ are falsely routed to another domain. This implies that we have 0.1% of the total traffic that are FRs. Given a large traffic, manually sifting through the entire traffic to find such utterances is prohibitively expensive and infeasible.

Previous work in Feedback-Based Self-Learning in Large-Scale Conversational AI Agents (Ponnusamy et al., 2020) uses implicit user feedback as a signal to generate automated reformulations. However, for low traffic domains, the low frequency of occurrence for utterances at the tail end of the distribution does not allow for automated reformulations at scale. Finetuning of pretrained language representation models which leverage inductive transfer learning (Howard and Ruder, 2018) have shown marked performance improvement with very small training datasets. Language models with bidirectional encoding such as BERT (Devlin et al., 2019; Liu et al., 2019) which are pretrained using a masked language modeling task have shown further improvements on the finetuning task due to its deep bidirectional encoder-based language representation. Further, larger models with 100s of billions of parameters like GPT3 (Brown et al., 2020) have demonstrated significant performance improvements with increase in model size. However, leveraging such models in production systems is challenging due to latency constraints.

We address this challenge by leveraging the power of BERT based pretrained language representation models in an offline setting. We combine utterance encodings from RoBERTa with embeddings of the N-Best hypothesis for the same utterance from the production system and conduct end-to-end finetuning in a multitask learning setting (Caruana, 1997) with a small, manually curated training set containing FRs. Multitask learning enhances language representation by discovering synergies between various finetuning tasks. We then use this model to sift through all the traffic and identify other FRs.

We tested our approach for identifying FRs from one domain in a large-scale conversational assistant. Compared to the baseline F1 score of 13% using a bi-LSTM model, our approach achieves an F1 of 25% using RoBERTa encodings. We further improve performance to 28% using N-best output, 30% using multitask learning, and to 32.3% using an ensemble of multitask models.

The rest of the paper is organized as follows: first, we describe the production NLU system; next we present details of our FR detection system and finally we present experimental results and conclusion.

## 2 Architecture of Large-Scale NLU Systems

Large-scale NLU systems are typically modularized into domain-specific modules where a domain is typically something like Books, Music, Shopping etc. As illustrated in Figure 1, each domain consists of several sub-components that are served by models for Domain, Intent and Named Entity Recognition. Let's consider an example user utterance U: "play hello by adele". Assume there are M domains. This utterance is sent to all of them and each domain then processes the utterance by sending it through its sub-modules described below.

(1) A Domain Classifier is a one-vs-all classifier

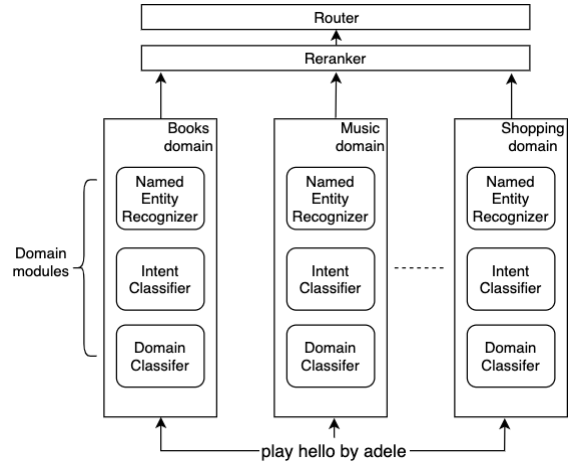

Figure 1: NLU System Architecture

indicating if the utterance belongs to that domain. For the example utterance U, a good classifier should output a high score for the positive class in the "Music" domain and output a high score for the negative class in all the other domains.

(2) An Intent Classifier is a multi-class model that provides intent specific scores for each of the intents belonging to a target domain. For the example utterance U, a good intent classifier should produce a high score for the intent "PlayMusic" when run in the "Music" domain.

(3) A Named Entity Recognizer (NER) that identifies named entities present in an utterance. For the utterance U, a good classifier in the "Music" domain should select "hello" and "adele" tokens as probable named entities.

To minimize run-time latency constraints, most large-scale production systems use a combination of deterministic artifacts (like rules) and simple models like MaxEnt (Berger et al., 1996) for these tasks. Once the domains process the utterance, the outputs are fed into a predictive model called reranker to obtain a sorted list. This sorted output is then used by the Router to send the utterance to the right domain for processing.

Due to the high volumes of traffic and the modular architecture with several domains, detecting the presence of errors (for instance, a domain misclassification) and isolating their sources is very challenging. Errors can occur and accumulate through every step of the processing flow highlighted in Figure 1. We next present our approach to identify one such error called False Reject (FR) where an utterance does not get routed to the correct domain by the production system.

## 3 False Reject Detection

For a given domain D, the task of identifying false rejects (FR) involves analyzing all traffic routed to a domain other than D and identifying utterances that should instead have been routed to D. The key challenges are the availability of only a small number of annotated examples of such FR and the extreme imbalance in our test set between FRs and utterances that are routed correctly.

### 3.1 Baseline System

As a baseline model, we used a bi-LSTM (Schuster and Paliwal, 1997) model to classify whether an utterance belongs to a domain D or not. Utterances tokenized into words and characters are leveraged in the form of their pre-trained Glove embeddings (Sakketou and Ampazis, 2020) and passed to a bi-LSTM layer. The stateful output of the bi-LSTM layer is passed through a fully-connected layer which outputs the probability of an utterance belonging to domain D as per equation (1) where $X_i$ represent the stateful output of the bi-LSTM layers, $\beta$ the learned weights of the fully-connected layer and $Y_i$ represents the classes.

$$\Pr(Y_i = 0) = \frac{e^{-\beta.X_i}}{1+e^{-\beta.X_i}} \quad (1)$$
$$\Pr(Y_i = 1) = 1 - \Pr(Y_i = 0) = \frac{1}{1+e^{-\beta.X_i}}$$

The training data used for fine-tuning comes from a low-cost human annotation effort where a small set of utterances are marked as "False Reject" (FR) or not by expert annotators. The False Rejects (FRs) for a domain D are then identified by applying the bi-LSTM based model on a test set of utterances that match the production traffic distribution. The utterances that are assigned to a domain D by the bi-LSTM based model but are not assigned to a domain D by the production model are filtered as the final FR candidates.

### 3.2 Pre-trained transformer models

Pre-trained transformer models such as BERT, GPT, T5 have proven to be quite effective at multitude of NLU tasks. Specifically, they can be adapted to a downstream task by fine-tuning on a small-sized dataset.

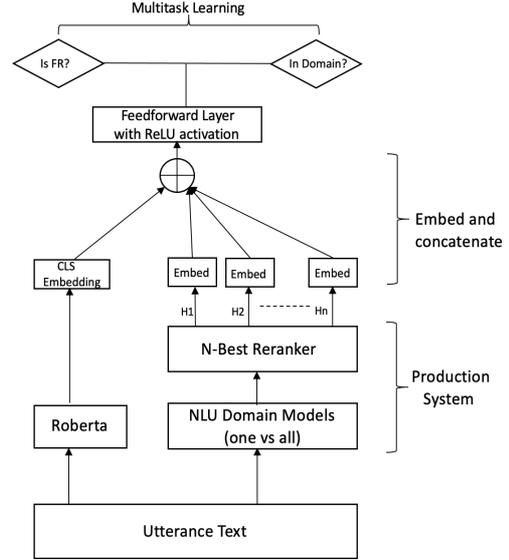

Figure 2: Transformer model leveraging N-Best from domain production models for FR detection

We leverage this strength for the task of FR detection. Specifically, we fine-tune a RoBERTa-based model for a binary classification problem in which given a domain D, we want the model to output the probability of an utterance belonging to that domain (represented as $\Pr(Y_i = 1)$ in equation (1). Instead of the bi-LSTM output, here $X_i$ represents the weights from the pre-trained model passed through a fully-connected layer. The model is trained to optimize the cross-entropy loss using the same dataset as the baseline system. We use the Adam optimizer with bias correction (Kingma and Ba, 2017) and a learning rate of $2e^{-5}$ with a warmup for 1/10th of the total number of training steps.

### 3.3 Leveraging N-Best Embeddings

The NLU system described in Section 3 produces an N-best hypothesis that represents the top N domains a user utterance likely belongs to. We tested incorporating this information into the FR detection model using two steps.

**Embed & Concatenate (N-best module)**: To leverage the N-best hypotheses, we choose top N hypotheses from the re-ranker, embed each of them into a 6-dimensional vector and concatenate them into a 6N-dimensional vector. This vector is further concatenated with the 1024 dimensional [CLS] token embedding from the pre-trained RoBERTa model. This (1024 + 6N) dimensional vector is then passed through a feed-forward layer to feed to the output layers.

**Multi-task output**: The model is trained to output predictions for two tasks. The first task is to determine if the utterance belongs to a domain or not. This is a generic task and is not specifically geared towards identifying if the production system falsely rejected an utterance. To this, we add a second classification task to specifically classify if the utterance is falsely rejected or not. These tasks could also share useful information across one another through their input hidden features.

The final model architecture that includes these components can be seen in Figure 2. A tokenized utterance is sent through the pre-trained RoBERTa model followed by the N-best module. The output from the N-best module is fed separately to two feed-forward layers to produce two outputs. The model is trained for 3 epochs using the same hyperparameters as the baseline transformer. The training took about 2 hours on a p3.dn24x large instance.

## 4 Results

We use a dataset with a FR to non-FR ratio of 1:15. This is not the true production distribution[1].

|  | Precision | Recall | F1 |
| --- | --- | --- | --- |
| bi-LSTM | 20.3 | 9.7 | 13.1 |
| RoBERTa | 50.1 | 16.3 | 25.2 |
| RoBERTa + N-Best single task model | 23.4 | 35.3 | 28.1 |
| RoBERTa + N-Best multitask model | 31.0 | 30.0 | 30.0 |
| RoBERTa + N-Best multitask model (ensemble) | **37.4** | **28.3** | **32.2** |

Table 1: Performance comparison of different FR detection techniques

However, it captures the imbalance and the low resource setting that we want to address in our framework. We hold out 15% of this dataset for validation. We have processed the data so that users are de-identified.

Table 1 compares the performance of different techniques we tested. We see a 12.1% absolute F1-score improvement from bi-LSTM to RoBERTa illustrating the value of pre-training for problems with extreme class imbalance and small training sets.

Incorporating the N-Best embeddings produces a further 3.1% improvement illustrating the value of using information from the production system itself to detect errors in its output. Using a multitask learning setting improves it further by 1.9% showing how similar tasks share learning representations. Ensembling produces an additional 2.2% improvement that is consistent with other machine learning results.

**Figure 3** presents a comparison of the precision-recall curves for RoBERTa and RoBERTa+N-Best multitask models. While the performance is comparable at high precisions, at lower values, the latter significantly outperforms. This is because the standalone RoBERTa model identifies a small number of frequent FRs, while the RoBERTa+N-Best multitask model is able to identify a greater variety of FR patterns without compromising precision significantly. Thus, the latter model offers a better tradeoff in identifying FRs that can then be used as data to train the production domain classifiers.

### 4.1 Production model improvements

Our RoBERTa-large based model is not directly usable in production due to latency constraints. As a result, we use a human-in-the-loop system where we leverage our model by running it in an offline setting to identify false reject candidates. These false reject candidates are sent for human annotation and the resulting annotations are fed as training data to the production model. We've noticed that the annotation throughput of true false rejects is improved by 5x due to the curated candidates provided by the offline transformer model. Furthermore, because these annotations are used to further train the production model, the False Rejects are reduced by about 20%.

---

[1] For confidentiality reasons, we don't use the true production distribution.

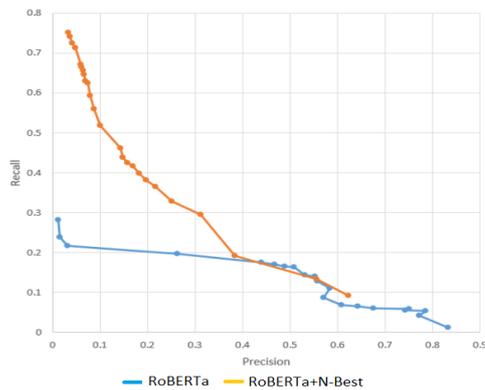

Figure 3: Precision-Recall Curve for RoBERTa and RoBERTa+N-Best embedding multitask

## 5 Conclusion

Our work presents a system for leveraging the power of computationally intensive, but accurate pre-trained language models to identify errors in a large-scale conversational assistant through offline analysis. Our results also demonstrate the effectiveness of such models in problems with large class imbalances. Specifically, we achieve an F1 score of 25.2% outperforming a bi-LSTM baseline by 12.1%. Further, we show that combining the output of the production system with pre-trained language models produce significant improvements (of 4.8 F1 points). As future work, we plan to leverage even larger models such as Megatron (Shoeybi et al., 2019) and T5 (Raffel et al., 2019) to achieve further improvements for the FR detection task.